# Sequential Thresholds:
# Context Sensitive Default Extensions

**Choh Man Teng**
teng@cs.rochester.edu
Department of Computer Science
University of Rochester
Rochester, NY 14627, USA

## Abstract

Default logic encounters some conceptual difficulties in representing common sense reasoning tasks. We argue that we should not try to formulate modular default rules that are presumed to work in all or most circumstances. We need to take into account the importance of the context which is continuously evolving during the reasoning process. Sequential thresholding is a quantitative counterpart of default logic which makes explicit the role context plays in the construction of a non-monotonic extension. We present a semantic characterization of generic non-monotonic reasoning, as well as the instantiations pertaining to default logic and sequential thresholding. This provides a link between the two mechanisms as well as a way to integrate the two that can be beneficial to both.

## 1 INTRODUCTION

Non-monotonic reasoning models the intuitive process of making non-deductive inferences in the face of certain supportive but not conclusive evidence. Default logic [Reiter, 1980] is one of the most well known formalisms of non-monotonic reasoning. Although default logic has many attractive features, it also encounters some conceptual difficulties in representing common sense reasoning tasks. In some cases, the most "intuitive" and direct formulation of a scenario can lead to unexpected interactions and unintended extensions, and we need to revise the formulation to some less intuitive form in order to arrive at the intended results. A number of variants of default logic have been proposed to circumvent some of these difficulties encountered by Reiter's default logic, but they too in some cases produce counter-intuitive results [Łukaszewicz, 1988; Brewka, 1991; Gelfond et al., 1991; Delgrande et al., 1994; Mikitiuk and Truszczyński, 1995].

Most of these approaches downplay an important aspect: that each reasoning step is not done in isolation. One implicit assumption of default logic and its variants seems to be that default reasoning can be characterized using simple rules that work equally well across all typical situations. A brute force method to deal with the unintuitive use of a default rule in some particular situation is to add a condition to the justification list to explicitly block the rule when that situation arises. This was first suggested in [Reiter and Criscuolo, 1981], where the use of semi-normal default rules instead of normal default rules was introduced.

We argue that we should not hope to write down default rules that are both simple and intuitive across many situations. A rule that appears to be intuitive and straightforward at first sight may have side effects and unintended applications in some circumstances. We need to take into account the context that is continuously evolving when we reason non-monotonically. After we have accepted a default conclusion, we take that conclusion as true, and it becomes part of the context with respect to which we evaluate the remaining default rules. Thus the context consists of the statements we take as true at the moment: the initial set of given "facts" $F$, *plus* the consequents of the default rules that have been applied up to this point.

Another way to handle the problem is to appeal to some external source of information not communicated in the default rules, such as using inheritance or priority hierarchies to prefer more specific or higher priority default rules [Touretzky, 1984; Horty et al., 1987], and the utilization of probability theory for determining acceptable default rules [Pearl, 1989; Neufeld et al., 1990; Bacchus et al., 1993]. Our approach of sequential thresholding also makes use of probability, in a way closer to the spirit of non-monotonic reasoning and in particular default logic.

The rest of the paper is organized as follows. Section 2 outlines the distinction between non-monotonic reasoning and Bayesian based probabilistic reasoning. Section 3 gives a brief summary of default logic, and discusses some of its conceptual difficulties and how they are related to the evolving context that is not accessible to the default logic mechanism. Section 4 introduces a probabilistic non-monotonic reasoning formalism we call sequential thresholding, and Section 5 provides a semantic account of non-monotonic reasoning in general and default logic and sequential thresholding in particular. Section 6 discusses how the two mechanisms can benefit from each other, and Section 7 concludes the discussion.

## 2  NON-MONOTONIC DEFAULTS, MONOTONIC PROBABILITIES

As pointed out in [Kyburg, 1996], there is a distinction between uncertain inferences and uncertain conclusions. In the former, the inference mechanism is uncertain, but the conclusions thus obtained are regarded as perfectly certain. There may be multiple extensions, each of which representing a possible scenario, but within each extension there is no question about the "truth" of the inferred conclusions. We may need to retract such conclusions in the face of new information, hence the non-monotonicity. Default logic and many other logical formalisms such as autoepistemic logic [Moore, 1985] and circumscription [McCarthy, 1980] fall into this category.

On the other hand, most of the probabilistic community falls into the latter category. The rules of probability employed, such as Bayes' rule, are perfectly deductive, and it is the conclusions, of the form $\Pr(x) = r$, which convey the uncertainty. New information does not invalidate the previous probability calculations, since the original probability values are conditioned on a string of events without the new piece of information, while with the update, we are conditioning on an additional piece of information. The two probability distributions are distinct as they are conditioned on distinct sets of events. Therefore such probability mechanisms are not non-monotonic.

An exception to this monotonic probabilistic trend is Kyburg's probabilistic acceptance [Kyburg, 1961], which advocates accepting the conclusions fully when their probabilities are deemed high enough. The work described in this paper is along the same lines. We focus on default logic as a concrete example of non-monotonic reasoning formalisms, and discuss how a link between default logic and probability can be beneficial to both. Below let us briefly summarize the preliminary terminology of default logic.

## 3  DEFAULT LOGIC [Reiter, 1980]

Let $\mathcal{L}$ be a standard propositional language, and $\mathcal{P}$ be the finite set of propositional constants in $\mathcal{L}$. We denote the provability operator by $\vdash$, and for any $S \subseteq \mathcal{L}$, we have $\mathbf{Th}(S) = \{\phi : S \vdash \phi\}$.

**Definition 1** *A default rule is an expression $\frac{\alpha : M\beta_1, \ldots, M\beta_n}{\gamma}$, where $\alpha, \beta_1, \ldots, \beta_n, \gamma \in \mathcal{L}$. We call $\alpha$ the* prerequisite, *$\beta_1, \ldots, \beta_n$ the* justifications, *and $\gamma$ the* consequent *of the default rule. A default rule is* normal *if it is of the form $\frac{\alpha : M\gamma}{\gamma}$, and* semi-normal *if it is of the form $\frac{\alpha : M\beta \wedge \gamma}{\gamma}$. A default theory $\Delta$ is an ordered pair $\langle D, F \rangle$, where $D$ is a set of default rules and $F$ (facts) $\subseteq \mathcal{L}$.*

Loosely speaking, a default rule $\frac{\alpha : M\beta_1, \ldots, M\beta_n}{\gamma}$ represents that if $\alpha$ is provable, and $\neg\beta_1, \ldots, \neg\beta_n$ is each not provable, then we by default assert that $\gamma$ is true. For a default theory $\Delta = \langle D, F \rangle$, the known facts constitute $F$, and a theory extended from $F$ by applying the default rules in $D$ is known as an *extension* of $\Delta$, defined as follows.

**Definition 2** *Let $\Delta = \langle D, F \rangle$ be a default theory over $\mathcal{L}$, and $E \subseteq \mathcal{L}$. $\Gamma(E)$ is the smallest set satisfying the following three properties. [1] $F \subseteq \Gamma(E)$, [2] $\Gamma(E) = \mathbf{Th}(\Gamma(E))$, and [3] for every default rule $\frac{\alpha : M\beta_1, \ldots, M\beta_n}{\gamma} \in D$, if (a) $\alpha \in \Gamma(E)$, and (b) $\neg\beta_1, \ldots, \neg\beta_n \notin E$, then $\gamma \in \Gamma(E)$. $E$ is an* extension *of $\Delta$ iff $E$ is a fixed point of $\Gamma$, that is, $E = \Gamma(E)$.*

### 3.1  CONCEPTUAL DIFFICULTIES

At first glance, a default rule is very easy to understand: if the prerequisite is true, and each of the justifications is possible, then infer by default the consequent. Reiter first argued [Reiter, 1980] that normal defaults are expressive enough for most common sense applications. However, on closer inspection, default rules that appear intuitively reasonable in isolation can give rise to unintuitive results when taken together [Reiter and Criscuolo, 1981; Lukaszewicz, 1985; Poole, 1989]. The facts and default rules may interact in unexpected ways, and result in no extension or unwanted multiple extensions.

We argue that this problem is to be expected in the default logic framework. Consider the following canonical normal default theory.

**Example 3** $\Delta = \langle D, F \rangle$, where $D = \{\frac{a : Mb}{b}, \frac{a' : M\neg b}{\neg b}\}$, and $F = \{a, a'\}$. There are two extensions, $E_1 = \mathbf{Th}(\{a, a', b\})$, and $E_2 = \mathbf{Th}(\{a, a', \neg b\})$.

Which extension is better? Should we get rid of one, or should we keep both? Let us consider the following popular instantiations.

- $a : \text{bird}, a' : \text{penguin}, b : \text{fly}$, we want only $E_2$.

- $a : \text{bird}, a' : \text{animal}, b : \text{fly}$, we want only $E_1$.

- $a : \text{quaker}, a' : \text{republican}, b : \text{pacifist}$, then we want to treat $E_1$ and $E_2$ equally.

The three instantiations are propositionally identical. Given the same default theory, we cannot expect the mechanism to generate different results. It is quite obvious that we, the "intelligent beings", are in possession of some contextual information that leads us to want to draw different conclusions in the different examples. However, this additional information, which underlies our "intuition" in choosing between the extensions, is not available to the default theory as it is formulated above. The theory is indifferent to whether $a'$ is supposed to stand for penguin or animal. Nor does it know that penguins are animals either. □

This problem motivates semi-normal defaults [Reiter and Criscuolo, 1981], where an extra justification is used to denote "non-exceptions", so that the rule can only be applied when the object in question is not a known exception. For example, the default rules in the first case discussed above would be amended as follows.

$$\{\frac{\text{bird} : \mathbf{M}\neg\text{penguin} \wedge \text{fly}}{\text{fly}}, \frac{\text{penguin} : \mathbf{M}\neg\text{fly}}{\neg\text{fly}}\}$$

where the term ¬penguin is added to the justification to prevent the rule from firing when the bird in question is a penguin.

This amended rule works fine in this case, but there are many other exceptions to the "birds typically fly" rule. This leads to a qualification problem, where we need to have a longer and longer $\beta$ in the justification part in order to cover all "unintuitive" cases. This seems to defeat the purpose of doing non-monotonic reasoning in the first place: we can as well explicitly list all exceptions, and revert to deductive inferences instead, which is not desirable even if feasible. "Naming defaults" [Poole, 1987] also has a similar problem.

### 3.2 MISSING CONTEXT

We argue that this "intuition problem" arises because we are misled into thinking that we can formulate default rules that are modular. By that we mean the default rules are expected to "do the right thing" regardless of what else is present in the environment. For example, the rule "birds typically fly" should fire when we only know that Tweety is a bird, but the same rule should not fire when we know in addition that Tweety is a penguin. The incorporation of exceptions into the justification is an attempt to encode some of this contextual information into the rule itself, but this makes the rules very cumbersome and difficult to formulate.

A much simpler approach is to take default rules as just one component of a non-monotonic reasoning system. We evaluate each default rule with respect to a contextual element which is encoded in some way external to the rule itself. An example of this approach is the use of inheritance and priority hierarchies [Touretzky, 1984; Horty et al., 1987] to determine the more specific or higher priority extension(s). For example, from the inheritance hierarchy we know that penguins are a subclass of birds, and the more specific rule concerning penguins is given priority over the more general rule concerning birds.

However, it is not always straightforward to establish a specificity hierarchy, especially when the classes concerned do not have a strict set inclusion relation. For example, consider a world in which not *all* penguins are birds, that is, penguins are typically birds, but there are some exceptions. It is then not obvious on what basis we can claim that penguins are more specific than birds.[1]

### 3.3 DYNAMICS OF THE CONTEXT

We pointed out that a default rule has to be evaluated relative to the context it is situated in. Now we also want to stress that this context is not a constant even if we only consider a single situation requiring default reasoning. Rather, the context is continuously evolving during the course of the reasoning process, as we draw more and more default conclusions.

One property that is present in many non-monotonic reasoning systems is that we can build upon the non-monotonic conclusions drawn earlier to draw yet more non-monotonic conclusions. In default logic this corresponds to the chaining of default rules. Every time a default rule is applied, the consequent $\gamma$ of that rule is added as part of the context available to the remaining rules, and $\gamma$ acquires virtually the same status as the statements in the original given set of facts $F$ of the default theory in subsequent computations. The default consequent of one rule can act as the prerequisite of another rule, and the latter rule is dependent on the former to create the favorable conditions for itself. Thus, even within one reasoning episode, the relevant

---

[1] Of course *all* penguins are birds, but consider replacing "birds", "fly", and "penguins" with "adults", "employed", and "university students" [Reiter and Criscuolo, 1981], and we would need to show that "university students" are more specific than "adults", a dubious assertion.

context is not a static component; rather each default rule operates within a different context.

One way to make this evolving context available to the reasoning machinery without explicitly and exhaustively listing all possibilities and conditions is to appeal to probability and the conditioning operation. Below we show a quantitative counterpart of default logic that retains its non-monotonic characteristic.

## 4 SEQUENTIAL THRESHOLDING

Now we present a probabilistic reasoning mechanism that is non-monotonic and very close in flavor to default logic. Sequential thresholding can be classified as a kind of probabilistic acceptance [Kyburg, 1961]. We accept a statement as true when its associated probability is high, say, above a certain threshold. The novelty here is that the thresholding is done sequentially, that is, each statement is considered in turn, and all accepted statements participate in the conditioning of subsequent statements in determining whether they are above threshold.

The thresholding process can be thought of as a form of generalized conditioning. In regular conditioning, the conditional probability $\Pr(\psi \mid \phi)$ is computed when $\Pr(\phi)$ is assumed to have changed to 1. Thresholding imposes a weaker requirement: given a threshold parameter $\epsilon$, accept $\phi$ if $\Pr(\phi) \geq 1 - \epsilon$, and adjust the posterior probabilities of all statements as if they are conditioned on $\phi$.[2]

More formally, we have the following definition.

**Definition 4** *A threshold collection $C$ is an ordered pair $\langle T, F \rangle$, where $T$ and $F \subseteq \mathcal{L}$. We call $T$ the threshold set and $F$ the fact set.*

*A filtered sequence $\Phi$ of $C = \langle T, F \rangle$ with respect to the sequential threshold parameter $\epsilon$ is a sequence $\langle \phi_1, \ldots, \phi_n \rangle$, such that [1] $\forall \phi_i$ in $\Phi$: $\phi_i \in T$ and $\Pr(\phi_i \mid F \cup \{\phi_1, \ldots, \phi_{i-1}\}) \geq 1 - \epsilon$, and [2] $\forall \phi \in T$ but not in $\Phi$, $\Pr(\phi \mid F \cup \{\phi_1, \ldots, \phi_n\}) < 1 - \epsilon$.*

$\Pr_\Phi(\psi)$, *the threshold probability of $\psi$ with respect to a filtered sequence $\Phi = \langle \phi_1, \ldots, \phi_n \rangle$ of $C$, is defined as $\Pr_\Phi(\psi) = \Pr(\psi \mid F \cup \{\phi_1, \ldots, \phi_n\})$.*

The filtered sequence $\Phi$ contains the formulas that we accept as true using the sequential thresholding process. The threshold probability $\Pr_\Phi(\psi)$ is the probability value of interest: the probability of $\psi$ conditioned on the facts $F$ and all the formulas in the filtered sequence, which are the formulas we have decided to accept.

The formulas in the fact set $F$ do not need to be above threshold to enter into the computation of the threshold probability. The formulas in the threshold set $T$, however, have to pass a threshold test in order to be included in the filtered sequence $\Phi$.

The value compared in the threshold test is the probability of $\phi_i$ conditioned on the facts $F$ and all the formulas that have been entered into the filtered sequence before $\phi_i$. This sequence is maximal, in the sense that no other formula in $T$ that is not already in $\Phi$ can be appended to the end of the sequence in accordance to the threshold test.

Note that the effective probability space is shrinking. If $\Pr(\phi_1 \mid F)$ is above threshold, we then only consider the space in which $F$ and $\phi_1$ are true in subsequent tests, and the probability distribution of interest becomes $\Pr(\star \mid F \cup \{\phi_1\})$, which becomes $\Pr(\star \mid F \cup \{\phi_1, \phi_2, \ldots \phi_k\})$ as more formulas are admitted into the filtered sequence. The probability value associated with a formula is technically unchanging: for example $\Pr(\psi \mid F \cup \{\phi_0, \phi_1\})$ is invariant throughout. Rather it is the "context" in which to compute the target probability value of $\psi$ that is continuously evolving, as we build the filtered sequence.

Note also the sequential nature of how the threshold test is administered. Since in general $\Pr(\psi \mid \phi_0, \ldots, \phi_k) \neq \Pr(\psi \mid \phi_0, \ldots, \phi_k, \phi_{k+1})$, a formula that is below threshold at one point might become eligible for thresholding after we have taken a few other formulas as true, and a formula that is above threshold at one point may not be so after we have thresholded some other formulas. In this sense, sequential thresholding is non-monotonic; we commit to the truth of certain formulas when they become above threshold, and the probability distribution of interest is changed accordingly.

## 5 DRAWING A PARALLEL

In this section, we characterize a generic non-monotonic reasoning mechanism in terms of the context that evolves as the computation progresses. This generic process can be instantiated in various ways to characterize various specific reasoning formalisms, in particular, default logic and sequential thresholding, thus providing a link between the two.

Informally, a non-monotonic reasoning step can be depicted generically as follows.

1. Find a non-monotonic rule that is applicable in the current context.

---

[2] The $\epsilon$ in the threshold does not tend to 0 in the limit, as in $\epsilon$-semantics [Adams, 1975; Pearl, 1989], but is assumed to be some small quantity.

2. Apply the rule, and accept its non-monotonic conclusion as true.

3. Extend the context to reflect the addition of the newly inferred conclusion.

After one cycle we find another rule that is applicable in the extended context, apply this rule, and add its non-monotonic conclusion to the new context, and the process repeats...

Both default logic and sequential thresholding conform to this characterization. For simplicity, in the case of default logic, we give the instantiation for normal default theories only. Non-normal default theories require some additional consideration, but the results basically hold for the general case as well [Teng, 1997].

Normal default theories have essentially the same structure as sequential thresholding. This can best be shown semantically by looking at the possible world partition sequences [Teng, 1996] of the generic non-monotonic reasoning mechanism described above, and of its instantiations for these two specific formalisms.

### 5.1 GENERIC NON-MONOTONIC PARTITION SEQUENCES

A possible world is characterized by a truth assignment to all propositions in the language, and a real number weight. Given an exhaustive set of possible worlds $W$, a partition sequence of $W$ is a tuple $\langle W_0, \ldots, W_l \rangle$, $l \geq 1$, where the non-empty sets $W_i$ in the sequence form a partition[3] of $W$.

**Definition 5** A generic non-monotonic reasoning partition sequence can be characterized by the process $\text{Partition}_W(\text{B}, \text{R})$, as follows.

**Input** An exhaustive set of possible worlds $W$, a set of background facts B, a set of non-monotonic rules R, each of the form $r = \langle \text{cond}; \text{res} \rangle$, where cond is the condition under which the rule $r$ can be applied, and res is the non-monotonic conclusion to be inferred if $r$ is indeed applied.

**Output** A partition sequence $\langle W_0, \ldots, W_l \rangle$ depicting the non-monotonic reasoning process.

**Process**

$W_0$: **Initialization** $W_0$ contains all the worlds in which the set of background facts B is false.

$W_i$: **Non-monotonic Step** For each $W_i$, $0 < i < l$, find a non-monotonic rule $r = \langle \text{cond}; \text{res} \rangle \in \text{R}$ such that

**Applicability** cond is satisfied in the context $(W_i, \ldots, W_l)$.
**Context Expansion** All the worlds in which res is false are grouped into $W_i$. The revised context of reference becomes $(W_{i+1}, \ldots, W_l)$.

$W_l$: **Closure** Every rule in R whose cond is satisfied in the context $(W_l)$ has been applied.

Given a set of facts B and a set of non-monotonic rules R, $\text{Partition}_W(\text{B}, \text{R})$ describes one way of how a partition sequence characterizing the reasoning with respect to B and R progresses. Each reasoning step is performed with respect to a *context* $(W_i, \ldots, W_l)$, which represents the body of information taken as true at the moment. The partition sequence can be seen as recording the change in this context as we draw more and more non-monotonic conclusions. The context evolves, starting at $(W_0, \ldots, W_l)$, and changing to $(W_1, \ldots, W_l)$, $\ldots$, $(W_i, \ldots, W_l)$, $\ldots$, until we reach $(W_l)$. Successive contexts differ by one formula res, which is the non-monotonic conclusion of the rule most recently applied.

Note that some of the $W_i$'s may be empty. For example, in the initialization step, if B is empty, then $W_0$ is empty. The last class $W_l$ in the partition sequence is of particular interest. It is the "end-state" of the reasoning process. In default logic, $W_l$ represents a default extension, and in sequential thresholding, this corresponds to the thresholded population. Also note that $\text{Partition}_W(\text{B}, \text{R})$ is not a function; there can be more than one partition sequence that can be constructed from the same initial parameters, such as when there are multiple extensions.

### 5.2 INSTANTIATIONS

Now we give the instantiations for normal default theory and sequential thresholding respectively. We borrow the notation of modal logic: $\Box \alpha$ is taken to mean $\alpha$ is true in all worlds, $\Diamond \gamma$ is taken to mean $\gamma$ is true in some world, both with respect to the reference context $(W_i, \ldots, W_l)$ in the following.

**Theorem 6** *Given a normal default theory $\Delta = \langle D, F \rangle$, a default partition sequence can be constructed by $\text{Partition}_W(F, \text{R})$, where*

$$\text{R} = \{\langle \Box \alpha \wedge \Diamond \gamma;\ \gamma \rangle : \frac{\alpha : \mathbf{M}\gamma}{\gamma} \in D\}.$$

*$E$ is an extension of $\Delta$ iff there is a default partition sequence $\langle W_0, \ldots, W_l \rangle$ such that $E$ is the set of sentences true in all the worlds in $W_l$.*

In the following, let $\%(\phi)$ denote the weighted proportion of worlds in which $\phi$ is true among the worlds in the context $(W_i, \ldots, W_l)$.

---
[3] A partition of a set $S$ is a set of *non-empty* sets $S_1, \ldots, S_n$, such that $\bigcup_i S_i = S$, and $S_i \cap S_j = \emptyset$ for $i \neq j$.

**Theorem 7** *Given a threshold collection $C = \langle T, F \rangle$ and a threshold parameter $\epsilon$, a sequential threshold partition sequence can be constructed by* $\text{Partition}_W(F, \text{R})$, *where*

$$\text{R} = \{\langle \%(\phi) \geq 1 - \epsilon;\ \phi \rangle : \phi \in T\}.$$

$\Phi = \langle \phi_1, \ldots, \phi_{l-1} \rangle$ *is a filtered sequence of $C$ at $\epsilon$ iff there is a sequential threshold partition sequence $\langle W_0, \ldots, W_l \rangle$ such that $r_i = \langle \%(\phi_i) \geq 1 - \epsilon;\ \phi_i \rangle$ is the rule used at step $i$. The threshold probability $\Pr_\Phi(\psi)$ is the weighted proportion of worlds in which $\psi$ is true in $W_l$.*

The partition sequences for normal default theories and sequential thresholding sport very similar conditions. The set of facts used to construct $W_0$ is given by a fact set $F$ in both formalisms. In normal default theory, a default rule $\frac{\alpha : M\gamma}{\gamma}$ can be applied when $\alpha$ is true and $\gamma$ is possible in the current context of $(W_i, \ldots, W_l)$. After this rule is applied, we take $\gamma$ as true, and exclude all those worlds in which $\gamma$ is false when we construct the revised context $(W_{i+1}, \ldots, W_l)$.

Similarly in sequential thresholding, $\phi_i \in T$ is appended to the filtered sequence when the proportion of worlds in which $\phi_i$ is true among those in the current context $(W_i, \ldots, W_l)$ is above threshold. After we have thresholded $\phi_i$, we take $\phi_i$ to be true, and exclude all those worlds in which $\phi_i$ is false when we construct the revised context $(W_{i+1}, \ldots, W_l)$.

For default logic, the last class $W_l$ determines what formulas are present in that particular extension, while for sequential thresholding, $W_l$ represents the formulas we have accepted (the filtered sequence) and the threshold probabilities with respect to these formulas.

## 6   SYMBIOSIS

Given the similarities regarding the structures of default logic and sequential thresholding, we can combine them easily into a hybrid formalism that makes use of non-monotonic rules whose applicability conditions involve both default logic style and probabilistic threshold style components. In this section we discuss how each formalism can benefit from the other in a hybrid arrangement.

### 6.1   FINDING THE GOOD EXTENSIONS

We saw that unlike many other probabilistic formalisms, sequential thresholding is non-monotonic, in that previously thresholded formulas can enter into subsequent computations as (defeasibly) true and not just highly probable statements.

The similarity in structure of sequential thresholding and default logic provides a link from "intuition" to probabilistically grounded rules, with respect to the specific context a default rule faces when it is applied during the reasoning process. This can be used to establish a context-sensitive metric for evaluating the relative "goodness" of multiple extensions, a situation often encountered in default logic.

An extension is constructed from a default theory $\Delta = \langle D, F \rangle$ by starting with the set of facts $F$, and applying a sequence of default rules $\langle d_1, \ldots, d_{l-1} \rangle$ successively. Let each default rule be normal and of the form $d_i = \frac{\alpha_i : M\gamma_i}{\gamma_i}$. The same partition sequence corresponding to this default extension can be achieved by sequential thresholding, using $F$ as the fact set and $\langle \gamma_1, \ldots, \gamma_{l-1} \rangle$ as the filtered sequence, with respect to some $\epsilon$ large enough so that all the $\gamma$'s can be above threshold.

There can be various ways to define a goodness measure of an extension as a function of this $\epsilon$ value. For example, one way is to take the minimum value $\epsilon_{\min}$ that would yield the extension in question. The sequential threshold $1 - \epsilon_{\min}$ measures how probable the most improbable (in the sequential context) default consequent is among those of the default rules used in constructing the particular extension. We can consider an extension "better" than another if its associated minimum threshold parameter $\epsilon_{\min}$ is smaller.

In contrast to other works relating default logic to probability, such as [Pearl, 1989; Neufeld et al., 1990; Goldszmidt et al., 1990; Bacchus et al., 1993], sequential thresholding retains the non-monotonic nature of default logic. This gives a fairer measure of the goodness of an extension, since it obeys a fundamental principle motivating default logic. In particular, both default logic and sequential thresholding rely on the evolving context resulting from the non-monotonic acceptance of conclusions.

### 6.2   WHAT'S IN IT FOR THRESHOLDING?

Some may argue the main strength of default logic over probabilistic approaches is that we can avoid having to come up with messy numbers. Why, one would then ask, do we need to have default rules at all when we have the probabilities?

People are not good at assigning numeric values to propositions. This difficulty alone provides a good incentive for seeking out formalisms that either do not make use of numeric information, or if they do, are not sensitive to perturbations in the numbers assigned. Sequential thresholding by itself is very sensitive to the value of the threshold parameter $\epsilon$ [Teng, 1996], and thus it makes more sense to use it as a companion mechanism for ranking default extensions rather than using it as a standalone reasoning mechanism. We are not as dependent on the actual numbers assigned when we only need to use them as a way to determine the relative goodness of the extensions.

Another problem of thresholding concerns the composition of the threshold set $T$, which contains the formulas we will attempt to threshold. Sequential thresholding does not deal with how we come up with this set. However, there is no reason to assume that we should threshold exhaustively until there are no more formulas above threshold (except tautologies), nor that it suffices to try arbitrary formulas. Until we have a reliable way of automatically generating candidates of "interesting and relevant" formulas, we need some mechanism for directing the thresholding process to focus on the formulas of our choice. Logical contraptions such as default logic provide a concise and well defined tool for expressing preferences in choosing the next formula to threshold.

### 6.3 EXAMPLE

An extended example is not possible here due to space limitation, but we will reuse the canonical Example 3 to give a sense of how the different formalisms compare.

A partition sequence corresponding to $E_1$ is $S_1 = \langle W_a, W_b, W_c \rangle$, where

$$\begin{aligned} W_a &= \{\{\neg a, \neg a', \neg b\}, \{\neg a, \neg a', b\}, \{\neg a, a', \neg b\}, \\ &\quad \{\neg a, a', b\}, \{a, \neg a', \neg b\}, \{a, \neg a', b\}\}, \\ W_b &= \{\{a, a', \neg b\}\}, \\ W_c &= \{\{a, a', b\}\}. \end{aligned}$$

A partition sequence corresponding to $E_2$ is $S_2 = \langle W_a, W_c, W_b \rangle$.

In sequential threshold terms, we have a threshold collection $\langle T, F \rangle$, where $T = \{b, \neg b\}$, and $F = \{a, a'\}$. The partition sequence $S_1$ characterizes the sequential thresholding process with respect to the filtered sequence $\langle b \rangle$, while $S_2$ can be generated with the filtered sequence $\langle \neg b \rangle$. The minimum $\epsilon$ that would yield $S_1$ is the weighted proportion of worlds in $W_b$ among those in both $W_b \cup W_c$. The minimum $\epsilon$ for $S_2$ is the weighted proportion of $W_c$ with respect to $W_b \cup W_c$.

Now consider the instantiation $a$ : bird, $a'$ : penguin, $b$ : fly. We know that flying penguins are quite rare, but how rare? The constraints on the magnitude of the weights assigned to the worlds in order to make $E_2$ more favorable than $E_1$ is very loose: we just need to make sure that the weight of $\{a, a', b\}$ (flying penguins) is smaller than the weight of $\{a, a', \neg b\}$ (non-flying penguins). This will ensure that the minimum $\epsilon$ required for $S_2$ is smaller than that for $S_1$.

It is very obvious in this toy example that the weights of the two worlds in question should differ greatly. However, in more complicated situations, we may only have a vague idea of the distribution of weights, and it would be helpful to know that there can be some leeway in the specification.

## 7  CONCLUSION

In non-monotonic reasoning, we accept certain conclusions as true although we do not know for sure that they are. All subsequent reasoning is carried out with the assumption that these accepted conclusions are indeed true. We argued that the problem of choosing between multiple extensions in default logic is due to the lack of contextual information. The evolving context during the course of the reasoning process is important in determining the "intuition" behind a default rule. This context can be made accessible in a few ways, one of which is sequential thresholding, a quantitative non-monotonic formalism that is very similar in spirit to default logic. Sequential thresholds can be used to evaluate the goodness of default extensions by accessing how probable each default rule is with respect to the continuously evolving context.

We presented a generic semantic characterization of non-monotonic reasoning, as well as the instantiations pertaining to normal default logic and sequential thresholding. This progressive addition of conclusions to the context, or the progressive restriction of admissible sets of possible worlds in $\langle W_i, \ldots, W_l \rangle$, is the crux of non-monotonic reasoning: the conclusions of the applied rules are taken as true, and the worlds in which these non-deductive conclusions are false are eliminated from further consideration.

We stress here that the rules are not modular, in the sense that they are not intended to be specified in a way that makes them independent of the rest of the rules and facts. On the contrary, the applicability condition cond of a rule depends critically on the current context, which is an environment created by the facts and the use of other rules. A more rigorous formalization of the generic non-monotonic partition sequences and further discussion can be found in [Teng, 1997].

### Acknowledgements

Many thanks to my advisor Henry Kyburg, for his many comments and repeated assurance that this is interesting, and for his NSF grant IRI-9411267.